\documentclass{article}
\usepackage{spconf,amsmath,graphicx}
 \usepackage{lingmacros}
 \usepackage{xcolor}
 
\title{A data-driven investigation of Noise-adaptive utterance \\ generation with Linguistic modification}

\name{Anupama Chingacham, Vera Demberg, Dietrich Klakow }
\address{Saarland Informatics Campus, Saarland University, Germany
}

\usepackage{tikz}

\newcommand\copyrighttext{%
  \footnotesize \textcopyright Copyright 2023 IEEE. Published in the 2022 IEEE Spoken Language Technology Workshop (SLT) (SLT 2022), scheduled for 9-12 January 2023 in Doha, Qatar. Personal use of this material is permitted. However, permission to reprint/republish this material for advertising or promotional purposes or for creating new collective works for resale or redistribution to servers or lists, or to reuse any copyrighted component of this work in other works, must be obtained from the IEEE. Contact: Manager, Copyrights and Permissions / IEEE Service Center / 445 Hoes Lane / P.O. Box 1331 / Piscataway, NJ 08855-1331, USA. Telephone: + Intl. 908-562-3966.}
\newcommand\newcopyrightnotice{%
\begin{tikzpicture}[remember picture,overlay]
\node[anchor=south,yshift=10pt] at (current page.south) {\fbox{\parbox{\dimexpr\textwidth-\fboxsep-\fboxrule\relax}{\copyrighttext}}};
\end{tikzpicture}%
}

\begin{document}

\maketitle
\newcopyrightnotice

\vspace{-5mm}
\begin{abstract}


In noisy environments, speech can be hard to understand for humans. 
Spoken dialog systems can help to enhance the intelligibility of their output, either by modifying the speech synthesis (e.g., imitate Lombard speech) or by optimizing the language generation.
We here focus on the second type of approach, by which an intended message is realized with words that are more intelligible in a specific noisy environment.
By conducting a speech perception experiment, we created a dataset of 900 paraphrases in babble noise, perceived by native English speakers with normal hearing.
We find that careful selection of paraphrases can improve intelligibility by 33\% at SNR \textbf{-}5 dB. Our analysis of the data shows that the intelligibility differences between paraphrases are mainly driven by noise-robust acoustic cues. Furthermore, we propose an intelligibility-aware paraphrase ranking model, which outperforms baseline models with a relative improvement of 31.37\% at SNR -5 dB. 

\end{abstract}

\begin{keywords}
noise-adaptive speech, paraphrases
\end{keywords}

\section{Introduction}

Over the past decade, speech-based interfaces have become an increasingly common mode of human-machine interaction.
Today, spoken dialog systems (SDS) are part of several systems such as those used for medical assistance, language learning, navigation and so on.
To improve the performance of SDS in the noisy conditions of daily-life, earlier studies have largely focused on speech enhancements for better automatic speech recognition (ASR).
But there is considerably less work on speech synthesis techniques to improve human recognition in noise.
However, to improve the human-like behaviours in SDS, speech synthesis needs to be adaptive to the noisy conditions.


Prior work has shown that \textit{acoustic modifications} can alter the intelligibility of speech uttered in an adverse listening condition~\cite{cooke2014listening}.
Synthesis of Lombard speech~\cite{hmmBasedLombard, LombardTransferLearning}, vowel space expansion~\cite{roger2012VowelSpaceTTS}, speech rate reduction and insertion of additional pauses~\cite{dall2014investigating} are some of the existing algorithmic solutions to reduce the noise impact on the intelligibility of synthesized speech. 
However, earlier studies have also highlighted the counter-productive effect of, signal distortions introduced by some noise-reduction techniques~\cite{cooke2014listening, brons2014effects}. 
On the other hand, \textit{linguistic modifications} are seldom leveraged by SDS to improve the utterance intelligibility, even though it is well-known that the speech perception in noise is significantly influenced by linguistic characteristics such as predictability~\cite{kalikow1977development, mansfield2021asr}, word familiarity~\cite{doi:10.1121/1.5090196}, neighborhood density~\cite{luce1998recognizing}, syntactic structure ~\cite{carroll2013effects, van2018evaluation}, word order ~\cite{uslar2013development} etc.
In particular, it was shown earlier that different types of noise, affect some speech sounds more than others~\cite{ mcardle2008predicting, vitevitch2002naturalistic, weber2003consonant, vanOs2021FrontPsych}. This opens the possibility to specifically choose lexical items that are less affected by the interference of a specific type of noise.

To this end, we propose an alternate strategy based on linguistic modifications to improve speech perception in noise. 
More precisely, we utilize the potential of \textit{sentential paraphrases} to represent the meaning of a message using lexical forms which exhibit better noise-robustness.  
One of the earlier approaches of utilizing linguistic forms to reduce word misperceptions in noise consisted of modeling phoneme confusions and pre-selecting less confusable words~\cite{cox2004modelling}.
Although their proposed model predicts the position of potential confusions in short phrases (which are formed by a closed vocabulary), the applicability of this approach for conversational data has not yet been studied.
Rational strategies like lexical/phrasal repetitions~\cite{cooke2014listening, kleijn2015optimizeSI} and insertion of clarification expressions~\cite{ogata2005generating} have also showed the possibility of improving the speech perception in noise without acoustic modifications. 
However, the scope of such template-based strategies are limited, as it may lead to the generation of less natural-sounding and monotonous utterances.
Compared to those earlier attempts, our work is more closely related to the study on rephrasing-based intelligibility enhancement~\cite{zhang2013rephrasing}. 
Zhang et al., 2013 focused on the development of an objective measure to distinguish phrases based on their intelligibility in noise.

In this paper, we concentrate on studying the impact of paraphrasing on utterance intelligibility at different levels of a noise type.
The current work is inspired by the earlier finding that
\textit{lexical intelligibility} in noise can be improved by replacing a word with its noise-robust synonym~\cite{chingacham21_interspeech}.
While this is an interesting finding, a sentence intelligibility improvement strategy solely based on lexical replacements is constrained by the availability of synonyms that fit the context of a given utterance. 
Hence, a paraphrase generation model was employed to include more generic types of sentential paraphrases.
Speech perception experiments were conducted at three different levels of babble noise: 5 dB, 0 dB, and -5 dB. 
We collected data from 90 native English speakers regarding their comprehension of 900 paraphrase pairs in the presence of babble noise. 
To date, this constitutes the largest available corpus of its kind.\footnote{Experiment data is released with an open-source license at: \\ https://github.com/SFB1102/A4-ParaphrasesinNoise.git}

Further, we investigated the influence of both linguistic and acoustic cues on intelligibility differences between paraphrases in noise. 
We utilized the speech intelligibility metric, STOI~\cite{taal2010short} to capture the amount of acoustic cues that survived the energetic masking, in a noise-contaminated utterance.
This metric also indicates the potential of listening-in-the-dips~\cite{cooke2006glimpsing}, as noise-robust acoustic cues capture the glimpses of the actual speech.
Additionally, a pre-trained language model was used for estimating the predictability offered by linguistic cues in an utterance.
Our modeling experiments reveal that the impact of paraphrasing on utterance intelligibility increases, as the noise level increases.
Also, we found that the observed gain in intelligibility is mainly introduced by paraphrases with noise-robust acoustic cues. 
For instance, consider the following paraphrases (\textit{s1},\textit{s2}), which are similar in linguistic predictability; yet distinct in intelligibility in the presence of babble noise:
\vspace{-2mm}
\begin{itemize}
    \itemsep0em 
    \item[\textit{s1}:]
    \textit{it never hurts to have some kind of a grounding in law.}
    \item[\textit{s2}:] 
    \textit{it doesn't hurt to have some kind of legal education.} 
\end{itemize}

\begin{figure}[t!]
    \centering
    \includegraphics[scale=0.24]{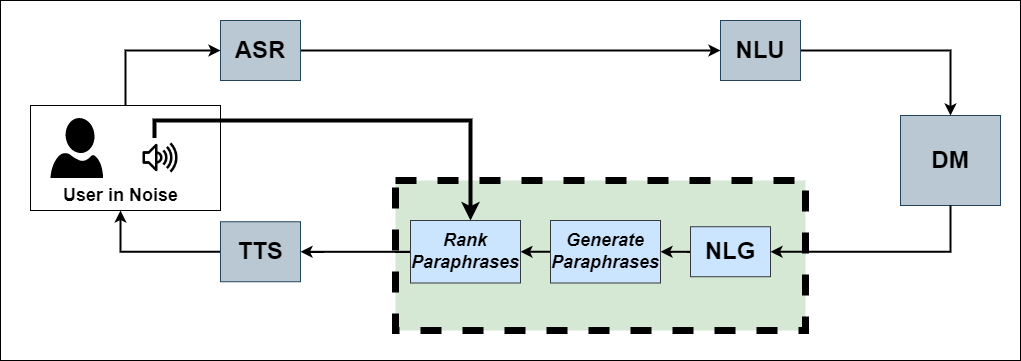}
    \vspace{-0.5em}
    \caption{Architecture of the proposed solution to generate noise-adaptive utterances using linguistic modifications.}
    \vspace{-1em}
    \label{fig:SDS_noise_adaptive}
\end{figure}

More concretely, at SNR~0 dB, listeners perceived \textit{s2} (1.2 times) better than \textit{s1}.
Further analysis of this distinction in intelligibility showed that, the acoustic cues which survived the energetic masking, is more in \textit{s2}, than in \textit{s1}.
Subsequently, our final step consists of demonstrating that it is possible to automatically predict among a pair of paraphrases, which of them will be more robust to noise (see Section~\ref{sec:svm_rank}). As shown in Figure~\ref{fig:SDS_noise_adaptive}, such ranking models could further be deployed in the language generation module of SDS, to generate noise-adaptive utterances without signal distortions.
Here, we assume that the noise in the user's environment is either known in prior or can be estimated.

\section{Sentential Paraphrases}
\label{sec:sent_para}

Paraphrases are those phrases/sentences which represent \textit{similar semantics} using different wording. However, the \mbox{notion} comes with the difficulty that two different sentences rarely have the exact same meaning in all contexts, hence paraphrases, especially at the sentence level \cite{bhagat2013paraphrase} typically only approximate the original meaning.
On the one hand, generating sentential paraphrases which are \textit{exactly equivalent} in semantics leads to trivial patterns such as word order changes or minimal lexical substitutions among paraphrases \cite{madnani2010generating}. This however can mean that there is only a minimal difference in the effect of intelligibility in noise between such paraphrases.
On the other hand, generation of non-trivial paraphrases introduces better lexical/syntactic diversity, and may hence have larger effects on intelligibility, but this in turn demands more scrutiny for semantic similarity \cite{dolan-brockett-2005-automatically}.

In this paper, we hence explore the effect of paraphrases that approximate semantic equivalence instead of strict semantic equivalence.
In order to include a large variety of paraphrases, stimuli sentences were generated using a pre-trained text generation model \cite{pegasus-paraphraser, zhang2020pegasus} which was fine-tuned on several paraphrase datasets like Quora Question Pairs, PAWS \cite{paws2019naacl} etc.
For the input sentences to the paraphrasing model, we selected a list of short sentences (10-12 words) from the dialogue corpus Switchboard~\cite{10.5555/1895550.1895693}.
After paraphrase generation, we employed automatic filtering to select the top two paraphrases for each input sentence, based on semantic similarity score\cite{zhang2019bertscore}. This resulted in a list of paraphrase triplets (\textit{s1, s2, s3}), consisting of different paraphrase types formed by lexical replacements, changes in syntactic structure etc. 
Since existing paraphrasing models lack the domain knowledge of spoken data, a manual selection was performed to ensure the quality of the generated paraphrases in terms of semantic equivalence. Every paraphrase triplet was converted to three pairs: (\textit{s1, s2}), (\textit{s2, s3}) and (\textit{s1, s3}).
Then, every paraphrase pair was verified for closeness in semantics. 
We identified about 300 triplets that exhibited approximate semantic similarity in all three pairs.
Those triplets were randomly split into three groups of 100 (one for each listening environment).
Hereafter, we refer to this dataset as \textit{paraphrases in noise} (PiN). Table~\ref{tab:sample_triplets} lists few samples in PiN.
\begin{table}[t!]
    \centering
    \begin{tabular}{ll}
    \textit{They seem to give more of just the facts than
     opinions.} \\
    \textit{They give more information than opinions.} \\
    \textit{They seem to give more facts than opinions.} \\
    
    \hline
    \textit{You never hear about it really in the big ones.} \\
    \textit{You don't hear much about it in the big ones.} \\
    \textit{In the big ones you don't hear about it.} \\
    
    \hline
    \textit{It was a very close game and hard fought game.} \\
    \textit{The game was close and hard fought.} \\
    \textit{It was a very close game.} \\

    \end{tabular}
    \caption{A few examples of (\textit{s1, s2, s3}) in the PiN dataset.}
    \vspace{-1em}
    \label{tab:sample_triplets}
\end{table}
To ensure that the sentential paraphrases in the PiN dataset are indeed equivalent in semantics, two annotators were asked to label the pairs as `paraphrase' only if they fit the definition of a `quasi-paraphrase', i.e., \textit{sentences or phrases that convey approximately the same meaning using different words}~\cite{bhagat2013paraphrase}.
Around 2/3 of PiN were identified as `\textit{quasi paraphrase}' by atleast one of the two annotators ($\textrm{PiN}_{either}$)  and 1/3 by both annotators ($\textrm{PiN}_{both}$).
Table \ref{tab:datasets_3} shows details of these subsets. 


\begin{table}[thb]
    \centering
    
    \begin{tabular}{lcccc}
         Dataset & Total & SNR~5 & SNR~0 & \mbox{SNR~-5} \\ 
         \hline
         PiN & 900 & 300 & 300 & 300 \\ 
         $\textrm{PiN}_{both}$ & 332 & 104 & 123 & 105 \\
         $\textrm{PiN}_{either}$ & 596 & 195 & 205 & 196 \\
    \end{tabular}
    \caption{Number of paraphrase pairs per listening condition. $\textrm{PiN}_{both}$ and $\textrm{PiN}_{either}$ are subsets of the PiN dataset.} 
    \vspace{-1em}
    \label{tab:datasets_3}
\end{table}
\vspace{-4mm}

\section{Sentence-level Intelligibility}
\label{sec:Sent-Int}
As the objective of the current study is to compare the perception of sentential paraphrases, a measure of intelligibility at sentence level is required. Thus, we defined a measure called \textit{sentence-level intelligibility} which captures the noise-robustness of an utterance. This measure is motivated from earlier work on \textit{slips of the ear}, where the proportion of listeners who misrecognized a word \textit{w1} as another word \textit{w2} is considered as the consistency of a confusion \cite{doi:10.1121/1.4967185}. Similarly, to measure how well a target utterance ($T$) is perceived in a listening condition, the mean recognition rate of an utterance among a set of listeners (with normal hearing thresholds) was calculated, as shown in (\ref{eq:Sent-Int}). For this purpose, we ensured that every stimulus utterance was listened to by a set of $n$ listeners. 

\vspace{-3mm}
\begin{equation}
\label{eq:Sent-Int}
    \textrm{Sent-Int} (T) = \frac{1} {n} \sum_{i=1}^{n} \textrm{Recog-Rate} (T,P_{i})
\end{equation}
\vspace{-3mm}

Further, the rate of recognition at each listening instance $i$, is calculated by 
comparing the phonemic transcripts of target ($T$) and perceived ($P_i$) utterances, as shown in (\ref{eq:Recog-Rate}). 
The phonemic transcripts $T^{ph}$ and $P^{ph}_i$ were generated using a Grapheme-to-Phoneme(G2P) converter~\cite{g2p-en} (see Table~\ref{tab:sample_Recog_Rate_calculation} for examples).
For string comparison, we used the Levenshtein distance (Lev), which calculates the minimum number of edits (i.e., deletions/substitutions/insertions of phonemes) required to change $T^{ph}$ into $P^{ph}_i$.
\vspace{-3mm}
\begin{equation}
\label{eq:Recog-Rate}
    \textrm{Recog-Rate} (T,P_i) = 1 - \frac{ \textrm{Lev}(T^{ph}, P^{ph}_i)}{\#phonemes_{T^{ph}}}
\end{equation}
\vspace{-3mm}

An equal cost (of 1) was assigned for all edit operations.
The phoneme recognition rate was then calculated by first normalizing the edit distance by the number of phonemes in target and then subtracting this value from 1. 
This ensures that the measure is not sensitive to the target utterance length.
The intelligibility score for each utterance ranges between 0 (completely unintelligible)  and 1 (completely intelligible). 

\begin{table*}[htb]
    \centering
    \begin{tabular}{ccc}
        Tgt/Prc & Transcripts & Recog-Rate \\

        \hline
         $T$ & \textit{you never hear about it really in the big ones} & \\
         $T^{ph}$ & \textit{Y UW - N EH V ER - HH IY R - AH B AW T - IH T - R IH L IY - IH N - DH AH - B IH G - W AH N Z} & - \\
         \hline
         $P_1$ & \textit{it really is the big one} & \\
         $P^{ph}_{1}$ & \textit{IH T - R IH L IY - IH Z - DH AH - B IH G - W AH N} & 0.5 \\  
         \hline
         $P_2$ & \textit{he went about it really in the big one} & \\
         $P^{ph}_{2}$ & \textit{HH IY - W EH N T - AH B AW T - IH T - R IH L IY - IH N - DH AH - B IH G - W AH N} & 0.7 \\
      
    \end{tabular}
    \vspace{-1em}
    \caption{The phonemic transcripts of a sample target and perceived utterances are reported with respective recognition rates. 
    }
    \vspace{-1em}
    \label{tab:sample_Recog_Rate_calculation}
\end{table*}

\section{Listening Experiment }
\label{sec:list_exp}


\vspace{-2mm}
To investigate whether paraphrases differ in intelligibility, we conducted a human listening experiment with the PiN dataset. 
\vspace{-2mm}

\textbf{Stimuli: } For audio stimuli creation, first, clean utterances for each sentence were synthesized using the speech synthesizer API of Google Translate \cite{gTTS}.
Then, corresponding noisy utterances were generated by performing additive noise-mixing using the babble noise from the NOISEX-92 database \cite{varga1993assessment} and the mixing tool audio-SNR~\cite{audio-SNR}.
To avoid priming effects, 15 stimuli lists with 60 sentences, were created while making sure none of the lists contained sentences that were paraphrases of each other. 

\textbf{Participants:}
The experiment was deployed on a crowd-sourcing platform, Prolific\footnote{https://www.prolific.co/} using LingoTurk framework \cite{pusse2016lingoturk}. It was conducted with a group of 90 participants who are native speakers of British English, based in UK.
The group consisted of 60 women, 29 men, and one non-binary person. 
The average age of the cohort is 32.4 ranging from 19 to 50. 
Our experiment was not accessible to individuals who reported to have hearing difficulties.

\textbf{Design and procedure:}
Every audio stimulus was presented to six different participants. Participants were asked to transcribe after listening to each utterance.
Every participant had 60 listening instances (20 per noise level). Similar to the perception experiment described in \cite{chingacham21_interspeech}, participants were instructed to use a placeholder (...) to mark cases of completely unintelligible words. They were also encouraged to guess, when necessary.

\textbf{Analysis:}
Experiment execution was followed by the intelligibility calculation (using Eq.~\ref{eq:Sent-Int}) for all 900 utterances. 
The overall intelligibility in a listening environment was measured by averaging the sentence-level intelligibility across all its utterances. 

Using this score, we conducted several analyses. 
First, we compared the overall intelligibility at different noise levels for all three sets of paraphrase pairs: the full dataset, and the two subsets of stricter paraphrase pairs which were annotated as such by atleast one or both the human judges.
Further, we calculated the (absolute) difference in intelligibility for each paraphrase pair, in order to infer whether a variation in the linguistic form of a message introduces a difference in intelligibility. The mean of this score represents the overall impact of paraphrasing on intelligibility, ranging from 0 (no effect) to 1 (maximum effect), see Figure \ref{fig:PPiN_diffSI_babblenoise_histogram} below.

\vspace{-3mm}
\subsection{Results and Discussion}
\label{ssec:list_exp_analysis}
\textbf{Full dataset:}
As expected, we observed a significant reduction in the overall intelligibility with an increase in the noise level: 0.97 (SNR~5), 0.94 (SNR~0) and 0.71 (\mbox{SNR~-5}). 
We noticed that listeners' ability to recognize utterances at SNR~0 is not as severely damaged as the \mbox{SNR~-5} condition. Reasons for this, besides the lower effect of sound masking, could be that listeners also understand the context better and have more cognitive capacity available for generating predictions, which in turn help them to recognize the words~\cite{kalikow1977development, schoof2015high}.



At SNR~5 and 0, most of the words were correctly recognized. 
Therefore, most paraphrase pairs exhibit only very small differences in intelligibility. 
By conducting one sample t-test on intelligibility differences between paraphrase pairs, we noticed a small effect of paraphrasing at SNR~5 ($\mu=0.039$, $p<0.001$).
Further, Welch two sample t-tests were performed to study the impact of paraphrasing at different levels of noise.
Our results showed that the mean intelligibility difference between paraphrases at SNR~0 ($\mu=0.059$, $p<0.001$) is significantly above that of SNR~5.
However, their scores being close to 0.0 indicates the limited impact of using paraphrases on intelligibility, under low noise levels.
But, at \mbox{SNR~-5}, we observed a significantly larger impact of paraphrasing on intelligibility ($\mu=0.198$, $p{<}0.05$) compared to that at a medium noise level, SNR~0.
This highlights that paraphrases are perceived with considerably different rates of recognition under highly noisy conditions.

\begin{figure}[thb]
    \centering
    \includegraphics[scale=0.5]{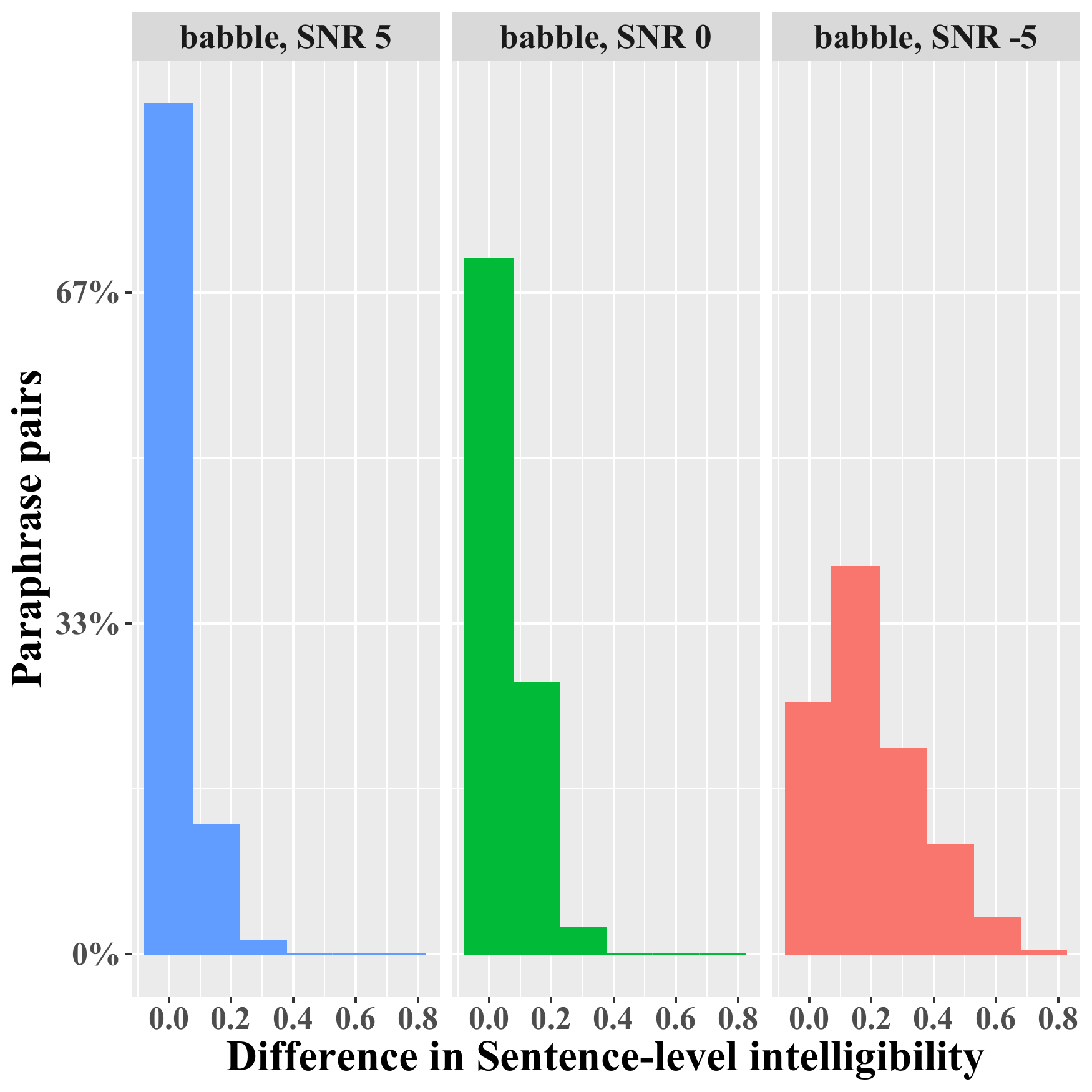}
     \vspace{-1em}
    \caption{
    The intelligibility difference between paraphrase pairs in the PiN dataset.
    The number of paraphrase pairs which are distinct in their intelligibility significantly increased with an increase in the \textit{babble noise} level.}
    \vspace{-1.5em}
    \label{fig:PPiN_diffSI_babblenoise_histogram}
\end{figure}


\textbf{Subsets of stricter paraphrases:}
Further, we analysed the mean intelligibility difference between the paraphrases in both subsets:  $\textrm{PiN}_{both}$ and $\textrm{PiN}_{either}$. 
Very similar to the PiN pairs, both subsets exhibited a steady increase in intelligibility differences from 0.035 at SNR~5, to 0.06 ($p{<}0.05$) at SNR~0 and 0.19  ($p{<}0.05$) at SNR $-5$.
This observation also aligns with an earlier finding that the recognition difference between synonyms increases with an increase in noise level~\cite{chingacham21_interspeech}. 
To measure the impact of paraphrasing on intelligibility, we considered an oracle model which correctly identifies the \textit{`more intelligible'} utterance (i.e. the one which has the highest Sent-Int) in every paraphrase pair.
By comparing the mean intelligibility of `more intelligible' paraphrases with `less intelligible' paraphrases at each noise condition, we observed a relative gain in overall intelligibility of 2\% at SNR~5, 5\% at SNR~0 and 33\% at \mbox{SNR~-5}. 
This leads us to the conclusion of this section that paraphrases can introduce differences in sentence-level intelligibility, suggesting the possibility of improving intelligibility by choosing a noise-robust paraphrase.

\vspace{-4mm}
\section{Modeling Experiments}
\label{sec:linear_regression}
Prior work has shown that speech perception in noise is influenced by both acoustic and linguistic characteristics.
However, there is little documentation in the literature to explain the features of paraphrases that introduce a gain in intelligibility, in noise. 
To this end, we conducted modeling experiments to study the impact of linguistic modifications on utterance intelligibility. 
For all our experiments, we utilized the implementation of linear regression models\footnote{code: https://github.com/SFB1102/A4-ParaphrasesinNoise.git/scripts} in the statistical software R~(Version 3.6.1) \cite{Rlanguage}.
We report the estimated model coefficients (\textbf{$\hat{\beta}$}) of individual features and their corresponding $p$-values.
A significance score of $p{<}0.05$ indicates the rejection of the null hypothesis ($H_0{: }\hat{\beta_j}{=}0$) that a predictor variable has no impact on the response variable.
The following three features were considered for defining the dependent variables:

\textbf{\textit{Length of utterance:}}
The length of utterance is an interesting correlate of intelligibility, as previous studies on speech modifications found that humans tend to shorten the utterance length for better perception in noise \cite{cooke2014listening}. 
While on the other hand, shorter words were found to be more confusing in noise \cite{luce1998recognizing}.
This feature is represented by the number of phonemes in an utterance (hereafter referred to as \textit{phLen}).

\textbf{\textit{Linguistic predictability:}}
Kalikow et al., 1977 showed that predictability of a word influences its intelligibility in noise.
The surprisal theory in language comprehension also demonstrates that the effort to process a word is inversely proportional to its predictability in context \cite{hale-2001-probabilistic}.
Similarly, earlier studies observed that listeners' perceptual difficulties are influenced by high level signals like linguistic predictability \cite{10.3389/fpsyg.2021.714485, coene2016linguistic} and situational cues \cite{ward2017effect}. 
However, predictability may also lead to false hearing instances \cite{rogers2012frequent}, where the listener is highly confident of the misheard utterance.
It is also interesting to study this feature in environments where the actual context of a word in a sentence is (acoustically) noisy and different from the linguistic context.
To represent how \textit{hard to predict} an utterance \textit{utt} $(u_0, u_1, u_2 ... u_t)$ is, we utilize the definition of perplexity, as stated in (\ref{eq:ppl}). 
\vspace{-3mm}
\begin{equation}
\label{eq:ppl}
    \textrm{PPL}(utt) = \exp { \{ - \frac{1}{t} \sum_{i=1}^{t}  \textrm{log} \ p_\theta(u_i|u_{<i}) \} }
\end{equation}
\vspace{-3mm}

Thus utterances which are less (linguistically) predictable are represented with high PPL scores.
For estimating the likelihood of a token from its preceding context, $p_\theta(u_i|u_{<i})$, a pre-trained dialog response generation model \cite{zhang2019dialogpt} was employed. Hereafter this feature is referred to as \textit{ppl}.

\textbf{\textit{Speech Intelligibility:}}
Speech Intelligibility (SI) metrics are widely used to perform speech enhancements and noise-reductions. 
The Short-Time Objective Intelligibility (STOI)~\cite{taal2010short} measure is one of the intrusive SI metrics, which requires the clean speech reference to estimate the intelligibility of a noisy speech.
The STOI value ranges between -1 and 1, as it captures the mean correlation between the time-frequency units of the clean and the distorted signal. 
Higher STOI values indicates better audibility.  \textit{STOI} scores were generated using a Python module \cite{pySTOI}.


\textbf{Analysis:} 
Prior to modeling the gain in intelligibility-induced by paraphrases, we studied the influence of above-listed features on \textit{sentence-level intelligibility} in noise. 
To this end, regression models were built separately for each noise level in the PiN dataset, by considering \textit{Sent-Int} as the response variable.
All models were fit to the data, after performing feature scaling with z-score normalization.
Similarly, for modelling the \textit{intelligibility-gain} in noise, we considered the absolute difference in \textit{Sent-Int} between paraphrase pairs (hereafter referred to as~\textit{Sent-Int-Gain}) as the response variable.
We hypothesize that the observed gain in sentence-level intelligibility can be explained by the relative difference in sentence-level features of paired paraphrases.
For this purpose, first, we identified the \textit{`more intelligible'} utterance in every paraphrase pair.
Then, we calculated the difference in features between paraphrase pairs, with respect to the \textit{`more intelligible'} utterance within each pair.
The predictor variables of this model are referred to as \textit{diff.phLen}, \textit{diff.ppl} and \textit{diff.STOI}.
In addition to PiN dataset, the two subsets of the dataset (as mentioned in Table~\ref{tab:datasets_3}) were also considered to model the intelligibility-gain among stricter paraphrase pairs. 

\vspace{-3mm}
\subsection{Results and Discussion:}
\vspace{-3mm}
\textbf{Sentence-level intelligibility:}
By considering the PiN pairs at SNR~5, we found that sentences which are shorter in length (\textit{phLen}: $\hat{\beta}$= -0.015, $p{<}0.05$) and linguistically more predictable (\textit{ppl}: $\hat{\beta}$= -0.007, $p{<}0.05$) are better perceived.
However, \textit{STOI} is insignificant at SNR~5, which highlights the negligible effect of a masker on perception, at a low noise level.
Moreover, we observed a main effect of acoustic cues, at higher noise levels like SNR~0 (\textit{STOI}: $\hat{\beta}$= 0.015, $p{<}0.05$) and \mbox{SNR~-5} (\textit{STOI}: $\hat{\beta}$= 0.070, $p{<}0.05$), indicating the relevance of noise-robust acoustic cues on speech perception.
Our models also showed a main effect of sentence predictability at higher noise levels: SNR~0 (\textit{ppl}: $\hat{\beta}$= -0.018, $p{<}0.05$) and \mbox{SNR~-5} (\textit{ppl}: $\hat{\beta}$= -0.049, $p{<}0.05$). 
This observation agrees with the earlier finding that linguistically predictable sentences are more intelligible in fluctuating noise condition\cite{schoof2015high}.
At \mbox{SNR~-5}, in addition to \textit{STOI} and \textit{ppl}, the model exhibited a main effect of \textit{phLen} ($\hat{\beta}$= -0.031, $p{<}0.05$), indicating better perception in noise with shorter utterances.
Overall, models indicate that stronger cues in top-down (linguistics) as well as bottom-up (acoustics) signals, lead to better utterance intelligibility in noise.
Next, we model the intelligibility-gain introduced by paraphrases in noise. 

\textbf{Gain in sentence-level intelligibility:} 
With stricter paraphrases in $\textrm{PiN}_{both}$ dataset, the model showed no impact of paraphrasing on intelligibility, at SNR~5.
This outcome is expected, as we observed a ceiling effect in sentence-level intelligibility at a low noise level (see Section~\ref{ssec:list_exp_analysis}).
However, the model exhibited a main effect of difference in acoustic-cues at both SNR~0 (\textit{diff.STOI}: $\hat{\beta}$= 0.014; $p{<}0.05$) and \mbox{SNR~-5} (\textit{diff.STOI}: $\hat{\beta}$= 0.066; $p{<}0.01$), indicating: (a) linguistic modifications can introduce utterances with better acoustic cues and (b) the observed intelligibility-gain at both SNR~0 
and \mbox{SNR~-5},  
is mainly driven by noise-robust acoustic cues.
By modeling the intelligibility-gain among $\textrm{PiN}_{either}$ pairs at SNR 5,
we found that \textit{diff.phLen} ($\hat{\beta}$= -0.008; $p{<}0.05$) exhibits, a small but a significant effect on \textit{Sent-Int-Gain}.
It indicates that paraphrases which are shorter in length than the given sentence, are better perceived in noise.
This observation also states the impact of approximation of semantic equivalence on the intelligibility-gain.
More precisely, compared to the $\textrm{PiN}_{either}$ subset, $\textrm{PiN}_{both}$ consists of more stricter paraphrases which are closer in semantics, but less different in their surface form.
Hence, the approximation in semantic equivalence among the $\textrm{PiN}_{either}$ pairs introduced a main effect of utterance length on intelligibility-gain. Similarly at \mbox{SNR~-5}, we observed the main effect of utterance length (\textit{phLen:} $\hat{\beta}$= -0.03; $p{<}0.05$) in addition to \textit{STOI} ($\hat{\beta}$= 0.04; $p{<}0.05$), representing that the gain in intelligibility being driven by both acoustic cues as well as the utterance length.  
However at SNR~0, we only found the main effect of STOI ($\hat{\beta}$= 0.009; $p{<}0.05$) on intelligibility-gain.
Like the $\textrm{PiN}_{either}$ pairs, the paraphrases in PiN exhibited similar effects on the intelligibility-gain, at SNR~5 and \mbox{SNR~-5}. 
However at SNR~0, we observed no statistically significant effect of \textit{STOI} with PiN dataset. This indicates the limitations of this acoustics-based metric to predict the intelligibility-gain at moderate noise levels.
In addition, we also noticed that the predictability of paraphrases showed no significant effect on the intelligibility-gain, under all three noise levels.
This observation is expected at SNR~5, as most of the utterances were perceived correctly.
However, the absence of predictability effect at higher noise levels indicates that the intelligibility-gain is less influenced by the difference in linguistic cues, introduced by the paraphrases in this dataset.
Overall, we found that the intelligibility-gain in noise is mainly driven by paraphrases with noise-robust acoustic cues. Additionally, shorter paraphrases also improved intelligibility in noise, however this was mostly observed among paraphrases which are less strict in semantic equivalence. 

\vspace{-3mm}
\section{Intelligibility-based Ranking Models}
\label{sec:svm_rank}

In order to automatically choose the \textit{more intelligible} utterance in a paraphrase pair, we trained an SVMRank model to perform pairwise ranking for each listening environment. 
For all our experiments, we utilized the model implementation by Joachims (2006).
Every model was trained with 80\% of the data and evaluated on the remaining 20\%.
Such pre-trained ranking models could further be utilized in dialogue generation systems to select the linguistic representations which are more robust to the noise in a listening environment.
Training models to rank sentences have already been explored in the past for different text processing tasks such as simplification \cite{vajjala-meurers-2014-assessing} and summarization \cite{madhuri2019extractive}. 
To quantify the performance of the ranking model, we used the percentage of correctly ranked pairs among the test set, referred to as \textit{pairwise ranking accuracy}.
All ranking models were trained and evaluated repeatedly with ten different train/test splits of the PiN dataset; their mean values are reported in Table~\ref{tab:svmrank_pin}.
We also performed an ablation study to quantify the influence of sentence-level feature(s) on ranking accuracy.

\textbf{Baselines}: For comparison, two baseline models, \textit{uniform} and \textit{majority} were considered. 
For each pair of paraphrases (\textit{s1, s2}), the predicted pair ranking has three options-- \textit{s1} is \textit{`more intelligible'} than \textit{s2}, \textit{s2} is \textit{`more intelligible'} than \textit{s1}, both \textit{s1} and \textit{s2} are equal in intelligibility. 
In the case of the \textit{uniform} baseline, an equal probability is given to all possible pair rankings which exist in the training set.
However for the \textit{majority} baseline, model always predicted the class which occurred the most in the training set.
At SNR 5 and 0, by sampling from a uniform distribution of all three ranking types, the \textit{uniform} baseline achieved an accuracy of $\sim$ 33\%.
However at \mbox{SNR~-5}, it achieved an accuracy of $\sim$ 50\%, as all paraphrase pairs differed in intelligibility.
Meanwhile the \textit{majority} baseline, performed equal to or better than the \textit{uniform} baseline.
By comparing the performance of ranking models with single features, we found that \textit{phLen} is a better feature than \textit{ppl} and \textit{STOI} at all noise levels.
Although \textit{phLen} achieves better performance than the baselines at SNR~5 and SNR~0, it fails to perform better than the \textit{uniform} baseline when the noise level is high. 
This indicates the necessity to consider other features for ranking. 
Considering the predictability as well, ranking improved by 5\% at both SNR~0 and \mbox{SNR~-5}.
However, including STOI in addition to \textit{ppl} and \textit{phLen}, further improved the ranking performance by 6\% at \mbox{SNR~-5}. 
In other words, this model achieved a relative improvement of 31.37\%, in comparison to the \textit{uniform} baseline.
This highlights the earlier observation of a significant effect of noise-robust acoustic cues on the intelligibility-gain among paraphrases.
Similarly, we observed this importance of STOI at \mbox{SNR~-5} being repeated for both subsets $\textrm{PiN}_{both}$ and $\textrm{PiN}_{either}$ by achieving a high accuracy of 70\% and 66\%  respectively.

\begin{table}[t!]
    \centering
    
    \begin{tabular}{llll}
        Feature(s) & SNR~5 & SNR~0 & \mbox{SNR~-5}  \\ 
        \hline 
        \textit{STOI}  & 46.0 +/- 2.5 & 49.0 +/- 3.7 & 53.0 +/- 3.1 \\ 
        \textit{ppl} &  39.0 +/- 2.5 & 52.0 +/- 3.1 & 55.0 +/- 3.1 \\ 
        \hline
        \textit{phLen} & \textbf{53.0 +/- 3.7*} & 59.0 +/- 3.7* & 56.0 +/- 3.7 \\
        + \textit{ppl} & 53.0 +/- 3.7* & \textbf{64.0 +/- 3.7*} & 61.0 +/- 3.1* \\
        + \textit{STOI} & 54.0 +/- 3.7* & 60.0 +/- 4.3* & \textbf{67.0 +/- 3.7*} \\ 
        \hline 
        \textit{majority} & 43.0 +/- 3.7 &  48.0 +/- 5.0 & 46.0 +/- 3.7 \\
        \textit{uniform} & 33.0 +/- 3.7 & 32.0 +/- 3.1 & 51.0 +/- 2.5

    \end{tabular}
    \vspace{-1em}
    \caption{
    The intelligibility based pairwise ranking accuracy of models built with the PiN dataset.
    Each score is a mean over ten runs (+/- 95\% CI). Scores with * are significantly better than both baselines. Bold-faced scores are the minimal models with considerably better accuracy at each noise level. 
    }
    \vspace{-1em}
    \label{tab:svmrank_pin}
    
\end{table}

\vspace{-2mm}

\section{Conclusion}

This work explored the possibilities of generating noise-adaptive utterances using sentential paraphrases.
In comparison to other noise reduction techniques, the proposed approach involves no signal distortion.
A dataset was collected to investigate the intelligibility differences between paraphrases, at three different levels of babble noise.
Our results from modeling experiments showed that paraphrases are capable to introduce an intelligibility-gain, as high as 33\%, at \mbox{SNR~-5} dB. 
In general, we found that the observed gain in intelligibility is mainly driven by paraphrases with more noise-robust acoustic cues. 
Our experiments demonstrated that with a proactive selection of the linguistic form to represent a message, better perception can be achieved in noise.
We believe the current contributions could further be extended to improve human-machine interactions in noise, especially for those who encounter noise-induced mishearing. 




\vspace{-2mm}

\section{Acknowledgements}
We thank Marjolein van Os, Dana Ruiter, and Iona Gessinger for proof reading and sharing their valuable feedback. 
This work is funded by the Deutsche Forschungsgemeinschaft (DFG, German Research Foundation) – Project-ID 232722074 – SFB 1102.

\bibliographystyle{IEEEtran}

\bibliography{sfba4_bib}

\end{document}